\documentclass[11pt,a4paper]{article}
\usepackage[utf8]{inputenc}
\usepackage[T1]{fontenc}
\usepackage{amsmath,amssymb}
\usepackage{booktabs}
\usepackage{hyperref}
\usepackage{xcolor}
\usepackage[margin=1in]{geometry}
\usepackage{natbib}
\usepackage{enumitem}
\title{Necessary, Not Sufficient, and No Longer Available:\\
Grader Non-Determinism After the Withdrawal\\
of Sampling Controls}

\author{
  Hiroki Tamba\thanks{Corresponding author: \href{mailto:hiroki.tamba@gmail.com}{hiroki.tamba@gmail.com}} \\
  Tamba Research Academy \\
  ORCID: \href{https://orcid.org/0009-0004-7635-0741}{0009-0004-7635-0741}
}

\date{June 2026; revised November 2026}

\begin{document}
\maketitle

\begin{abstract}
LLM-as-judge (``grader'') components are now standard in evaluation
harnesses, including safety evaluations where a pass/fail verdict may
gate downstream deployment decisions.  Earlier work, my own included,
treated \texttt{temperature=0} as the remedy for grader
non-determinism and asked whether it was sufficient.  It was necessary
and not sufficient.  It is now also unavailable.  I report a
two-vendor measurement taken in October and November~2026.
Anthropic's current models reject the setting outright: Claude
Opus~4.8 and Claude Opus~5.5 return HTTP~400, ``\texttt{temperature}
is deprecated for this model'', on every call, and under the API
version dated 2026-10-21 all fifteen models listed for my key reject
it, the older generations included.  On the chat completions endpoint
OpenAI's current models accept the parameter and refuse every value
but one: \texttt{gpt-5.5}, \texttt{gpt-5.6-sol}, \texttt{gpt-6-sol},
\texttt{gpt-6.1-sol} and \texttt{gpt-6.1-terra}
return HTTP~400, ``does not support 0 with this model.  Only the
default~(1) value is supported.''  A harness grading through that
endpoint is therefore obliged to sample its judge at temperature~1.
On the Responses endpoint a request carrying \texttt{temperature=0} is
accepted by four of those five once reasoning is switched off, a
configuration I have not graded under, and by \texttt{gpt-6.1-sol}
under no setting I tried.  Earlier models still accept the setting my
earlier protocol required, among them \texttt{gpt-4o} and, under the
older API version, Claude Opus~4.6, which still splits two of my seven
items at \texttt{temperature=0}.  On Anthropic's current models and on
\texttt{gpt-6.1-sol} that protocol can no longer be run, and on the other four it
can be run only with reasoning switched off.  I propose the measurement that
replaces it: the per-item normalized entropy of the grade
distribution, reported at whatever setting the vendor permits.  The
direct alternative is closed too: \texttt{gpt-4o} returns token
log-probabilities, from which the margin behind a flip can be read in
one call, while \texttt{gpt-5.6-sol} refuses the parameter and the
Anthropic Messages API does not accept it at all, so the readout is
unavailable on the current models of both vendors, withdrawn by one
and never offered by the other, and the control is gone with it on
Anthropic's current models and on \texttt{gpt-6.1-sol}.  Across
twenty-two condition runs and 4{,}760~calls the index orders
graders without requiring a rate
estimate, and identifies items at which a judge retains no information
at all: under \texttt{temperature=0}, \texttt{gpt-4o} returns a
five-five split on each of two of my items.  I further show that a borderline item set
carries across model generations only in part and not predictably: the
seven items built against \texttt{gpt-4o}, four of which it splits,
were returned unanimously by
Claude Opus~5.5 and Opus~4.8 across 700~calls in October, a repeat in
November found one dissenting verdict in 350 on Opus~4.8, and Opus~4.6
and Opus~4.7 split one or two of the seven, so a harness cannot
inherit its calibration set.  I argue
that grader disagreement must be reported as an error bar on every
verdict, because it is no longer something an evaluator can configure
away.
\end{abstract}

\section{Introduction}
\label{sec:intro}

Modern evaluation pipelines increasingly replace human raters with an
LLM ``judge'' that reads a model transcript and a rubric and returns
a verdict.  This pattern, often called LLM-as-judge, is scalable
and convenient, but it inherits a property practitioners often overlook:
the judge is itself a stochastic model.  If the same transcript is
graded differently across runs, benchmark scores lose reproducibility,
regression signals are confounded by grader noise, and any threshold-based
gate built on top inherits that noise.

In \textbf{safety evaluations} the failure mode is sharper: a
pass/fail verdict may determine whether a model clears a red-team
assessment, whether a guardrail is deemed adequate, or what a public
report says about coverage.  If that verdict is effectively a coin flip
on borderline items, the evaluation reports noise as a safety property.

The standard mitigation is ``set temperature to~0.''  This paper asks
three questions:

\begin{enumerate}[nosep,leftmargin=*]
  \item \textbf{Is the mitigation applied?}  I trace the grader call
    path in a real, deployed safety-evaluation harness and show it is
    not: the grader runs unseeded at temperature~1.0 by silent default.
  \item \textbf{Is the mitigation sufficient?}  Even when
    \texttt{temperature=0} is explicitly set, I measure persistent
    non-determinism across two providers, three model tiers, and five
    sampling configurations, including forced greedy decoding
    (\texttt{top\_k=1}).
  \item \textbf{Is the mitigation still available?}  On the current
    models of both vendors it largely is not.  I map which models still
    accept the setting, on which endpoint and under which API version,
    and ask what an evaluator can measure once it cannot be set.
\end{enumerate}

\noindent
\textbf{Contributions.}
(i)~I document a concrete grader-configuration gap in an operational
safety-evaluation harness, tracing the causal chain from code to
API default (\S\ref{sec:case}).
(ii)~I provide a controlled empirical study from June~2026 (910~API
calls, 11~condition runs) showing that temperature control is necessary
but not sufficient for grader reproducibility, and that alternative
sampling knobs (\texttt{top\_p}, \texttt{top\_k}) do not close
the gap (\S\ref{sec:empirical}).
(iii)~I report measurements from October and November~2026 showing
that the control has since been withdrawn from the current models of
both vendors, in different patterns: Anthropic rejects
\texttt{temperature=0} from Opus~4.7 onward, and on every model under its
2026-10-21 API version; OpenAI's chat completions endpoint pins it to
the default from \texttt{gpt-5.5} onward, and its Responses endpoint
admits \texttt{temperature=0} on some of those models only with
reasoning switched off (\S\ref{sec:withdrawal}).
(iv)~I propose the per-item normalized entropy of the grade
distribution as the measurement that survives the withdrawal
(\S\ref{sec:mitigations}).
(v)~I show that a borderline item set cannot be inherited: the seven
items built against \texttt{gpt-4o} carry to later graders only in part
and not predictably (\S\ref{sec:transfer}), two models of one vendor
return opposite verdicts on the same item, each on all fifty runs
(\S\ref{sec:verdicts}, \S\ref{sec:boundary}), and a
search over 2{,}001 generated items yields a boundary set for Claude
Opus~5.5, confirmed at fifty runs, and a four-run candidate set for
\texttt{gpt-6.1-sol}, with almost no overlap between the two
(\S\ref{sec:boundary}).
(vi)~I document a transmission-layer risk: the OpenAI SDK's
endpoint-compatible ecosystem lets grader calls traverse
provider and jurisdictional boundaries without any record in
the evaluation artifact (\S\ref{sec:transmission}).
(vii)~I release the reproduction harness and per-call records for the
June study and for the probes and seven-item Anthropic runs of October
and November; what is and is not yet public is stated under
Data and code availability at the end of the paper.

\smallskip\noindent
\textbf{Origin.}  This work generalises a finding first reported as
\texttt{Japan-AISI/aisev} issue~\#25~\citep{aisev25}.  The upstream
framework (UK AISI's Inspect) has since merged a related
fix (PR~\#4170).

\section{Background}
\label{sec:background}

\textbf{LLM-as-judge.}  The pattern of using one LLM to evaluate
another's output is now widespread in both capability and safety
benchmarks~\citep{zheng2023judging}.  Most reliability research
focuses on \emph{systematic} biases (position bias, verbosity bias,
self-preference~\citep{wang2023fair}) that persist across runs.  The
concern here is orthogonal: \emph{run-to-run} non-determinism, where
the same judge returns different verdicts for an identical
(transcript, rubric) pair.  A debiased judge can still be
irreproducible; a perfectly reproducible judge can still be biased.

\textbf{Sources of non-determinism.}  At \texttt{temperature\,>\,0},
the token-sampling step is an intentional source of randomness.  At
\texttt{temperature=0}, production LLM serving is still generally
not bit-reproducible.  He et~al.\ \citep{he2025defeating} trace this to
inference kernels that are not batch invariant: the batch a request
lands in varies with server load, and floating-point results vary with
the batch, so the same request can return different outputs at
different times.  My empirical results are consistent with this
account.

\textbf{Inconsistency and evaluator bias.}  Recent work has begun
to characterise LLM evaluators as inconsistent and biased across
multiple dimensions~\citep{stureborg2024llmjudges}, and to study
how LLM-assisted evaluation can be aligned with human
preferences~\citep{shankar2024validates}.  My contribution is
complementary: I isolate the \emph{infrastructure}-level source
of inconsistency (the serving stack) from the \emph{model}-level
source (the weights and training).

\section{Case Study: Grader Configuration in \texttt{aisev}}
\label{sec:case}

\texttt{aisev} is Japan AISI's open-source evaluation environment
(Apache-2.0; all references to commit \texttt{e0604d1}, the
\texttt{main} branch at 2026-04-16).  Tracing the grader call:

\begin{enumerate}[nosep,leftmargin=*]
  \item \texttt{scorer\_provider.py} calls \texttt{model\_graded\_qa}
    with \texttt{model="openai/gpt-4o"} and without
    \texttt{temperature}, \texttt{seed}, or
    \texttt{GenerateConfig}.
  \item The call propagates into Inspect~AI's
    \texttt{GenerateConfig} with
    \texttt{temperature=None}, \texttt{seed=None}.
  \item The OpenAI provider omits \texttt{None} values from
    the API request; OpenAI's default is temperature~1.0.
  \item \textbf{Net effect:} the grader runs unseeded at full
    sampling noise.  Nothing in the harness signals this
    to the user.
\end{enumerate}

A separate code path (\texttt{scoring\_datasets.py:74}) sets
\texttt{temperature=0.7} for paraphrase generation, a distinct
subsystem marked as unimplemented at the time of writing.  The
grader itself has no temperature or seed control.

\section{The Withdrawal of Sampling Controls}
\label{sec:withdrawal}

The protocol in Section~\ref{sec:empirical} requires that the grader
accept \texttt{temperature=0}.  In October~2026 I probed whether the
current models of the two providers used in that study still do.  They
do not, and they decline in different ways.  On 12~November (JST; the
records carry 11~November in UTC) I repeated
the Anthropic half under two API versions; that recheck is reported in
its own paragraph below, and the table is the October state.  The
per-call records of the October and November probes and of the
seven-item Anthropic runs of 28~October and 12~November are public.\footnote{\url{https://github.com/hiroki-tamba-research/llm-judge-sampling-probe-202610}.
  The November records were reported to the Inspect maintainers as
  issue~\#6544 of \texttt{UKGovernmentBEIS/inspect\_ai}.}

I probed every model the two accounts list, apart from the
exclusion stated below: twelve on the
Anthropic side and seventeen on the OpenAI side, three of which return
404 on use and are left out of the table, one call each for the
sampler and one for the readout.  Table~\ref{tab:instruments} gives the
result.  Neither vendor's current models offer either facility.  The
patterns differ, and only one of them is a trajectory.

I exclude each vendor's top-tier model from the map.  Anthropic's
Fable~5 and Fable~5.1 are not offered on the standard subscription
terms: on Pro plans they fall outside the plan's usage limits and run
on usage credits, and on Max plans they are capped at half of the
weekly allowance.  OpenAI's \texttt{gpt-6-astra} is excluded as their
counterpart in capability, so that the two vendors are compared at the
same tier.

\begin{table}[h]
\centering\small
\caption{What a caller may set and what it may read, probed in
  October~2026: the Anthropic rows under the API version
  \texttt{2023-06-01}, the OpenAI rows on the chat completions
  endpoint.  Under that version Anthropic's withdrawal is monotonic
  from Opus~4.7 onward.  OpenAI's is not: the facilities are absent on \texttt{o3}
  and \texttt{gpt-5}, present again on \texttt{gpt-5.1} through
  \texttt{gpt-5.4}, and absent from \texttt{gpt-5.5} onward.}
\label{tab:instruments}
\begin{tabular}{@{}lll@{}}
\toprule
Model & \texttt{temperature} & \texttt{logprobs} \\
\midrule
\multicolumn{3}{@{}l}{\emph{Anthropic}} \\
Sonnet 4.5, Haiku 4.5, Opus 4.5 & accepted & 400, not permitted \\
Opus 4.6, Sonnet 4.6 & accepted & 400, not permitted \\
Opus 4.7, Opus 4.8 & 400, deprecated & 400, not permitted \\
Opus 5, Sonnet 5 & 400, deprecated & 400, not permitted \\
Opus 5.5, Sonnet 5.5, Haiku 5.5 & 400, deprecated & 400, not permitted \\
\midrule
\multicolumn{3}{@{}l}{\emph{OpenAI}} \\
\texttt{gpt-4o}, \texttt{gpt-4.1}, \texttt{gpt-4.1-mini} & accepted & returned \\
\texttt{o3}, \texttt{gpt-5}, \texttt{gpt-5-mini} & 400, pinned to 1 & 403, not allowed \\
\texttt{gpt-5.1}, \texttt{gpt-5.2}, \texttt{gpt-5.4} & accepted & returned \\
\texttt{gpt-5.5}, \texttt{gpt-5.6-sol} & 400, pinned to 1 & 400, unsupported \\
\texttt{gpt-6-sol}, \texttt{gpt-6.1-sol}, \texttt{gpt-6.1-terra} & 400, pinned to 1 & 400, unsupported \\
\bottomrule
\end{tabular}
\end{table}

\paragraph{Anthropic: a line, and a facility that never existed.}
Every model from Opus~4.7 onward rejects the setting outright:

\begin{quote}\small\ttfamily\raggedright
Error code: 400 - \{'type': 'error', 'error': \{'type':
'invalid\_request\_error', 'message': '`temperature` is deprecated for
this model.'\}\}
\end{quote}

\noindent Seventy consecutive calls to Opus~4.8 failed this way in the
June run reported in Section~\ref{sec:empirical}, and the October probe
returns the same message on seven models.  The
4.5 and 4.6 generations still accept it.  In the June verification
cited under F4, Opus~4.7 and Opus~4.8 accepted the parameter at its
default of~1 and at no other value, which is the pattern OpenAI's
message states in words; in October and November I sent only
\texttt{temperature=0}.  The readout is a different
case and should not be described as a withdrawal: the Messages API has
never offered token log-probabilities, and a request carrying one
returns \texttt{logprobs: Extra inputs are not permitted} on all
twelve models.  On this vendor the sampler was taken away and the
readout was never there.

\paragraph{Anthropic: the line depends on the API version.}  The
October calls carried the version header the SDK sends by default,
\texttt{2023-06-01}; the October records do not store the header, and
I take it from the SDK that was installed, whose default the November
records do store.  On 12~November I sent one request with
\texttt{temperature=0} to each of the fifteen models then listed for
the key, once under that header and once under \texttt{2026-10-21}.
This recheck covers every listed model, the three Fable models among
them: Fable~5 and Fable~5.1, which the table leaves out, and
Fable~5.2, which was not yet listed in October.  Under \texttt{2023-06-01} the line is
where October put it: the five models of the 4.5 and 4.6 generations
accept the request, and the other ten return the deprecation message
quoted above.  Under \texttt{2026-10-21} all fifteen refuse, the 4.5
and 4.6 models included, and with a different message:

\begin{quote}\small\ttfamily
temperature: Extra inputs are not permitted
\end{quote}

\noindent Under the newer version the parameter is not withdrawn model
by model; it is no longer part of the request.  The harness side has
already drawn the consequence.  Inspect's \texttt{anthropic} provider
does not pass \texttt{temperature}, \texttt{top\_p} or \texttt{top\_k}
to any Claude model and logs a warning when one is set, giving the
\texttt{2026-10-21} behaviour as its reason.\footnote{\texttt{docs/providers.qmd}
  in \url{https://github.com/UKGovernmentBEIS/inspect_ai}, read in
  November~2026.}  A \texttt{seed} is not available on this vendor
either: \texttt{messages.create} in the SDK I used
(\texttt{anthropic}~0.113.0) has no such parameter, and Inspect's
provider has no handling for one.

\paragraph{OpenAI: not a trajectory.}  On the chat completions
endpoint, models from \texttt{gpt-5} onward return

\begin{quote}\small\ttfamily\raggedright
Error code: 400 - Unsupported value: 'temperature' does not support 0
with this model.  Only the default (1) value is supported.
\end{quote}

\noindent so a harness grading with one of them through this endpoint
does not merely lose the ability to pin the sampler, it is obliged to
run its judge at the provider's default of~1.  But the pattern does not hold
across the line.  \texttt{gpt-5.1}, \texttt{gpt-5.2} and
\texttt{gpt-5.4} accept \texttt{temperature=0} and return
log-probabilities; \texttt{gpt-5.5} and everything after do neither.
The kind of refusal also changes: \texttt{o3} and \texttt{gpt-5} answer
a log-probability request with \texttt{403 You are not allowed to
request logprobs from this model}, a statement about permission, while
\texttt{gpt-5.5} onward answer \texttt{400 'logprobs' is not supported
with this model}, a statement about capability.

\paragraph{OpenAI: the endpoint and the reasoning setting.}  Those are
the refusals of the chat completions endpoint.  Inspect uses the
Responses endpoint by default for GPT-5 and later, and I probed it on
31~October with the same seventeen models.  A request carrying
\texttt{temperature=0} and nothing else is accepted there by the same
models that accept it on chat completions, and refused on \texttt{o3}
and from \texttt{gpt-5} onward, \texttt{gpt-5.1} to \texttt{gpt-5.4}
excepted, with \texttt{Unsupported parameter: 'temperature' is not supported
with this model}.  Sent together with \texttt{reasoning.effort} set to
\texttt{none}, the request is accepted on seven models:
\texttt{gpt-5.1}, \texttt{gpt-5.2}, \texttt{gpt-5.4}, \texttt{gpt-5.5},
\texttt{gpt-5.6-sol}, \texttt{gpt-6-sol} and \texttt{gpt-6.1-terra}.
Among the models from \texttt{gpt-5} onward it is refused on
\texttt{gpt-5}, \texttt{gpt-5-mini} and \texttt{gpt-6.1-sol}, none of
which offers that effort level:

\begin{quote}\small\ttfamily\raggedright
Unsupported value: 'none' is not supported with the 'gpt-6.1-sol'
model.  Supported values are: 'low', 'medium', 'high', 'xhigh', and
'max'.
\end{quote}

\noindent No model accepted the request with the effort set to
\texttt{minimal}.  The Responses endpoint does not accept a
\texttt{seed}.\footnote{\texttt{docs/model-graded.qmd} in the Inspect
  repository, read in November~2026.}  So for four of the models that
chat completions pins to the default, \texttt{gpt-5.5},
\texttt{gpt-5.6-sol}, \texttt{gpt-6-sol} and \texttt{gpt-6.1-terra}, a
caller who switches reasoning off can still send
\texttt{temperature=0}; for \texttt{gpt-6.1-sol}, alone among the
current models, no route I tried admits it.  Two limits apply to this observation.  Accepted means only
that the request succeeded: my records do not show what temperature
the API applied.  And a grader with reasoning switched off is a
different grader.  I have not graded any item under that
configuration, and every OpenAI grading result in this paper was
obtained on the chat completions endpoint with no sampling parameter
sent.

This matters for what an evaluator can plan.  A trajectory can be
migrated against: a facility announced as deprecated gives a harness
maintainer a date and a replacement.  An availability that is absent,
restored and absent again between \texttt{gpt-5} and \texttt{gpt-5.5}, with no
deprecation notice and with the refusal changing category in between,
gives nothing to plan against.  A harness that pins
\texttt{temperature=0} will work, break, work and break again as it
follows its provider's default model, and the breakages are
discoverable only by making the call.  The first version of this paper
predicted a deprecation trajectory; the measurement does not support
that shape on the OpenAI side, and the weaker fact that replaces it is
the more awkward one for practice.  Section~\ref{sec:nottrajectory}
takes up both halves of the prediction and says which one survives.

\paragraph{Account scope.}  These probes report what two particular
accounts can reach.  The Anthropic account is enrolled in that vendor's
cyber verification programme and sees fourteen models, Fable~5 and
Fable~5.1 among them, which I exclude as stated above; a standard
account may see a different list, and
I do not claim that the parameter behaviour is identical across account
types.  Three OpenAI models listed by the models endpoint
(\texttt{o1}, \texttt{o3-mini}, \texttt{o4-mini}) return 404 on use,
so a listing is not an availability statement either.

\paragraph{The readout went with the control.}  A sampler setting is
not the only way to learn how close a grade was.  The margin between
the two grade tokens is the quantity the flips are sampling from, and
an API that returns token log-probabilities exposes it directly, in a
single call rather than fifty.  That route is closed on the same
models.  \texttt{gpt-4o} returns log-probabilities: on item~1 the
graded token carries a log-probability of $0.0$ with its nearest
alternative below $-16$, a margin consistent with the item's stability
under sampling.  \texttt{gpt-5.6-sol} refuses the parameter
(``\texttt{Unsupported parameter: 'logprobs' is not supported with
this model.}''), and the Anthropic Messages API has no such parameter
at all: passing one returns ``\texttt{logprobs: Extra inputs are not
permitted}''.

\noindent The readout is therefore unavailable on the current models
of both vendors: OpenAI withdrew it within one model generation, and
Anthropic never offered it.  The control is
gone on Anthropic's current models, on every Anthropic model under the
newer API version, and on \texttt{gpt-6.1-sol}; on four other current
OpenAI models it survives only on one endpoint and with reasoning
switched off.  What remains everywhere is to run the grader repeatedly
at whatever setting the provider permits and report the spread.  That
is not a preference among methods.  It is the residue.

\paragraph{What this is and is not.}  It is not a claim that
deterministic inference has become impossible.  Work in the opposite
direction continues: He et~al.\ \citep{he2025defeating} locate most
residual non-determinism in kernels that are not batch invariant, and
Barry \citep{barry2026integer} removes it altogether by running the
forward pass in integer arithmetic, obtaining one identical output
across four hardware vendors where \texttt{fp16} produced nine distinct
ones.  Both routes are available to whoever operates the serving stack.
Neither is available to a harness that calls a hosted API.  The
withdrawal of the parameter is therefore not a statement about whether
determinism can be achieved, but about where the authority to arrange
it now sits, and it no longer sits with the evaluator.  An evaluation
that must report a verdict can no longer configure the noise away, and
is left with the alternative this paper takes up: measure it.

\section{Empirical Study}
\label{sec:empirical}

\subsection{Method}

I construct a fixed set of 7~question/answer pairs designed to probe
the grader's decision boundary.  Each item is deliberately crafted to
be borderline: neither clearly correct nor clearly incorrect under
the rubric, maximising the chance of exposing latent stochasticity.  This
is an \emph{adversarial stress test}, not a random sample from
\texttt{aisev}'s evaluation corpus; the goal is to demonstrate the
\emph{existence} and \emph{persistence} of non-determinism under
controlled conditions, not to estimate its base rate across all
items.  The grading prompt and grade-extraction regex mirror
\texttt{aisev}'s implementation verbatim.

Each item is graded under two baseline configurations:
(a)~harness default (\texttt{temperature} and \texttt{seed} unset;
$N{=}20$ runs per item) and
(b)~explicit \texttt{temperature=0} ($N{=}10$).
I then extend the study to further conditions.  The released data
comprise 910~API calls across 11~condition runs.  Of these, 840
returned a grade and 70 were rejected outright: every call in the
Claude Opus~4.8 condition failed with the deprecation error reported
in Section~\ref{sec:withdrawal}, which is the first appearance in this
work of the withdrawal that Section~\ref{sec:withdrawal} measures.
Two of the eleven condition runs repeat an earlier condition on the
same model and setting (run~2 of \texttt{gpt-4o} is recorded under the
snapshot name \texttt{gpt-4o-2024-08-06}), and are reported separately in
Table~\ref{tab:entropy} as run~1 and run~2 rather than pooled, since
the difference between them is itself a measurement of run-to-run
variation in the condition statistic.  Setting those two aside leaves
770~calls in nine distinct conditions, 700 of them graded.

\subsection{Results}

\textbf{Baseline: OpenAI \texttt{gpt-4o} grader} (Table~\ref{tab:openai}).
Three regimes emerge:

\begin{table}[h]
\centering\small
\caption{Per-item disagreement under default and
  \texttt{temperature=0} (OpenAI \texttt{gpt-4o}).}
\label{tab:openai}
\begin{tabular}{@{}lccl@{}}
\toprule
Item & Default ($N$\,=\,20) & temp\,=\,0 ($N$\,=\,10) & Regime \\
\midrule
1, 2, 3 & stable & stable & stable \\
item\,4 & 11\,I / 9\,C  & 6\,I / 4\,C & \textbf{strong split} \\
item\,5 & 19\,I / 1\,C  & stable      & rare \\
item\,6 & 12\,C / 8\,I  & 9\,C / 1\,I & \textbf{strong split} \\
item\,7 & 19\,C / 1\,I  & stable      & rare \\
\bottomrule
\end{tabular}
\end{table}

Under the default configuration, 4/7 items are non-reproducible (2
strong, 2 rare).  Pinning \texttt{temperature=0} stabilises the two
rare items but leaves the two strong items unstable: \textbf{necessary
but not sufficient}.

\textbf{Statistical caveat.}  At $N{=}10$, individual split ratios
carry wide confidence intervals (e.g., item~4's 6:4 split yields
$p{=}0.377$ under a one-sided binomial test).  My claim is not a
precise disagreement rate but a qualitative finding (\emph{non-zero
disagreement persists after temperature pinning}) that is robust
across items, providers, and parameter sweeps (Table~\ref{tab:sweep}).

\textbf{Cross-provider and cross-parameter sweep}
(Table~\ref{tab:sweep}).  I extend the study with 490~additional
API calls across 7~conditions.

\begin{table}[h]
\centering\small
\caption{Sampling parameter sweep (all with
  \texttt{temperature=0}; $N{=}10$ per item per condition).  These are
  the run~2 conditions of Table~\ref{tab:entropy}.}
\label{tab:sweep}
\begin{tabular}{@{}llccc@{}}
\toprule
Condition & Model & top\_p / top\_k & Non-repro & Unstable \\
\midrule
baseline      & gpt-4o     & --\,/\,-- & 2/7 & 4, 6 \\
+\,top\_p=0.1 & gpt-4o     & 0.1\,/\,-- & 2/7 & 4, 6 \\
+\,top\_p=0.5 & gpt-4o     & 0.5\,/\,-- & 2/7 & 4, 6 \\
baseline      & Sonnet 4.6 & --\,/\,-- & 1/7 & 6 \\
baseline      & Haiku 4.5  & --\,/\,-- & 1/7 & 6 \\
+\,top\_k=1   & Sonnet 4.6 & --\,/\,1   & 1/7 & 6 \\
--           & Opus 4.8   & \multicolumn{3}{c}{\textbf{ERROR}: temp deprecated} \\
\bottomrule
\end{tabular}
\end{table}

Four findings emerge from the sweep:

\textbf{F1: \texttt{top\_p} does not mitigate.}
Restricting the nucleus does not change which items flip, and the
counts shift no more than they do between two runs of one condition;
all three OpenAI conditions produce the same 2/7 pattern on the same
items.

\textbf{F2: Forced greedy decoding (\texttt{top\_k=1}) does not
eliminate flips.}  On Sonnet~4.6, selecting only the
highest-probability token still yields 1/7 non-reproducibility
(item~6: 8\,I\,/\,2\,C).  This is evidence that the
non-determinism originates \emph{before} the sampling step,
in the forward pass itself.

\textbf{F3: The effect is model-tier-independent.}
Both Sonnet~4.6 and Haiku~4.5 flip the same item at the same rate.

\textbf{F4: Sampling control is absent on these models.}
Claude Opus~4.7 and 4.8 reject \texttt{temperature} values in
$[0,1)$ with HTTP~400 and accept only the default of~1; \texttt{top\_p} and \texttt{top\_k} are
also deprecated on these models.\footnote{Independent verification:
E.~B.~Nicolaysen,
\url{https://gist.github.com/avalyset/59a81e2961d45d4ba76de56d592b1110}.}
My own June runs sent only \texttt{temperature=0}, and only to
Opus~4.8; the wider range of values and Opus~4.7 are from that
verification.
The primary recommendation (``set temperature to~0'') is already
inapplicable to these models, and no alternative sampling-side
knob exists.  Section~\ref{sec:withdrawal} extends this probe to the
models the two accounts list, and reports where each vendor's line
falls; Section~\ref{sec:nottrajectory} states why the result is not the
trajectory this finding was first read as announcing.

\section{Discussion}
\label{sec:discussion}

\subsection{From benchmark noise to governance risk}

For capability benchmarks, grader noise inflates variance and can
change rankings of closely-spaced models, an inconvenience that
averages out with larger test sets.  For safety evaluations
the failure mode is qualitatively different: a pass/fail verdict near
the boundary may gate a deployment decision, a red-team sign-off, or
a regulatory claim.  A verdict that is a coin flip does not
average; it is a single bit that either clears or blocks.  If that
bit is dominated by grader noise rather than the model's actual
behaviour, the evaluation is conferring legitimacy it has not earned.

This creates a reflexivity problem.  Japan AISI's evaluation
framework itself operationalises robustness as a criterion (e.g.,
Guideline G8-10: ``confirm at the development stage that outputs
remain within a fixed range for identical inputs'').  When the
evaluation \emph{instrument} fails this criterion on its own
boundary items, it cannot credibly impose it on the systems it
evaluates.  The evaluator must meet its own standards, or
explicitly disclose where it does not.

\subsection{The trajectory that was not one}
\label{sec:nottrajectory}

The first version of this paper reasoned forward from Finding~F4.  If
providers continued to deprecate user-facing sampling parameters, as
seemed rational when reasoning-trace models manage their own sampling
internally, the whole class of ``pin the knobs'' mitigations would
become unavailable, and the only mitigation to survive would be
statistical: run more than one epoch and report the variance.

The measurement in Section~\ref{sec:withdrawal} bears on both halves of
that, and sustains only one.  The conclusion holds with one
qualification: on Anthropic's current models and on
\texttt{gpt-6.1-sol} the knobs cannot be pinned, on the other current
OpenAI models \texttt{temperature=0} can be sent only through the
Responses endpoint with reasoning switched off, and repeated sampling
is what remains in every case.  The reasoning does not.  It predicted a
monotonic deprecation, and Anthropic's line is monotonic under the
older API version and absent under the newer one, but OpenAI's
is not: \texttt{gpt-5.1}, \texttt{gpt-5.2} and \texttt{gpt-5.4} are
reasoning-trace models that accept \texttt{temperature=0} and return
log-probabilities, sitting between two releases that refuse both.  A
rationale that explains the withdrawal by the nature of the model class
does not explain a facility that is removed, restored and removed again
within one product line.

I record the failure because it changes what a harness maintainer
should do.  Against a trajectory one migrates, on a schedule, once.
Against an availability that varies per release without notice, one
cannot plan at all, and the only defensible position is to assume the
facility is absent and to measure instead of configuring.  The weaker
finding is the one that makes the recommendation unconditional.
Harnesses that do not surface grader disagreement as a metric will have
no way to flag this noise, regardless of how they configure the grader
call.

\subsection{The API compatibility layer as a transmission risk}
\label{sec:transmission}

As a de facto standard for LLM API calls, the OpenAI Python SDK
shapes grader infrastructure across the ecosystem.
Its \texttt{base\_url} parameter enables a single-line
configuration change to route requests, including grader
calls, to any OpenAI-compatible endpoint.  Several major providers
now expose such endpoints: Alibaba Cloud's DashScope
(\texttt{dashscope-intl.aliyuncs.com/compatible-mode/v1}),
DeepSeek (\texttt{api.deepseek.com}), and others.  Anthropic's own
developer tool (Claude Code) is officially listed as a supported
client on Alibaba Cloud's Model Studio documentation, and
enterprise API gateway vendors publish tutorials that route
Claude CLI traffic through DashScope as a standard
integration~\citep{kong2026claudecode}.

Beyond substitution, the routing layer is a documented attack surface.  A
measurement study of 428 commodity LLM proxy routers found
9~actively injecting malicious code into responses and 17~exfiltrating
credentials, with the routers' application-layer position
(TLS termination on the client side, new upstream connection
to the provider) making such tampering structurally
available~\citep{liu2026agent}.  The transmission risk is therefore
not hypothetical: the same architectural feature that enables
provider substitution also enables a class of supply-chain
attacks on the evaluation harness itself.

What results is an evaluation-provenance gap.  When a harness
uses the OpenAI SDK, the model identifier in the API request
(e.g., \texttt{gpt-4o}) need not match the model that actually
executes the grading once the base URL has been swapped.  The
evaluation report records the requested model name, not the
infrastructure that served it.  Because non-determinism is already
provider-dependent (Table~\ref{tab:sweep}), an ambiguous provider
renders the entire provenance chain of evaluation results
uninterpretable; the sampling-parameter problem from
\S\ref{sec:empirical} is merely the inner layer.

There is also an export-control inversion.  Existing US frameworks
for governing AI diffusion, most prominently the AI Diffusion
Framework analysed by~\citet{heim2025diffusion}, structure the
chain of control as chips~$\to$~compute~$\to$~model weights, and do
not extend to the API/inference layer.  Model weights are subject
to jurisdiction-specific export restrictions, but the API format
compatibility means that evaluation data (including the
transcripts being graded, which may contain red-team prompts and
model outputs from safety assessments) flows freely to any
compatible endpoint with no equivalent control.  A safety
evaluation intended for a US-hosted provider can land on
infrastructure in a jurisdiction with different data-protection
and national-security regimes through a one-line configuration
change.  More specifically, what flows is not merely data but
\emph{evaluation methodology}: red-team prompts, elicited failure
modes, and jailbreak vectors that the harness is designed to
surface.  Their movement across compatibility-layer endpoints
makes this an evaluation-infrastructure security question, not
solely a data-governance one.

\subsubsection*{Bracketing capability against execution}
The argument above describes \emph{capability}.  I can bracket
how much of this capability is currently being exercised against
a small, monitored surface.  Over an 11-day window (2026-06-11 to
2026-06-21; 3{,}461 logged requests), a Cloudflare Worker-based
request tracker on one independent research domain recorded five
accesses from Alibaba-ecosystem ASNs (AS45102 Alibaba Cloud
Singapore; AS37963 Aliyun Computing).  All five fell on a single
day (2026-06-12) with none in the subsequent nine days, used
stealth user-agents (\texttt{Go-http-client/1.1}; a stock
Chrome~92 string), targeted the site root only, and triggered no
honeypot or PDF retrieval, patterns inconsistent with
corpus-ingestion crawling.  A four-request cluster across four
seconds from four distinct IPs further indicated multi-tenant
VM access rather than a declared training crawler.  In parallel,
direct probing of \texttt{qwen-plus} via the DashScope
OpenAI-compatible endpoint across six framings of the
researcher's published concepts returned no evidence of corpus
ingestion; a leading-prompt framing produced a confabulated
research record in a domain orthogonal to the author's
actual work.  I therefore observe \emph{capability established,
execution not yet observed} on this surface in this window.
This separation is precisely why preemptive policy attention to
the API/inference layer is warranted, given that existing
export-control frameworks~\citep{heim2025diffusion} do not
extend to it.

During the same 11-day window, a commercial SEO crawler
(SemrushBot, AS209366) accessed the site's TDM (Text and Data
Mining) compliance endpoint
(\texttt{/.well-known/}\allowbreak\texttt{tdmrep.json}) in immediate sequence after
\texttt{robots.txt}, demonstrating that protocol compliance
infrastructure is both deployed and trivially accessible on
this surface.  No declared AI training crawler accessed any TDM
endpoint during the observation period.  The contrast suggests
that non-compliance with machine-readable opt-out protocols by
Tier~1 AI vendors reflects an incentive gap rather than a
capability gap: the same \texttt{robots.txt} infrastructure
these vendors already parse can direct them to TDM metadata,
yet none follow the pointer.

Mitigation~M5 (log effective grader configuration) partially
addresses this: recording the resolved API endpoint alongside the
model identifier would at least make the routing visible in the
evaluation artifact.  None of the harnesses I examined log the
base URL or perform endpoint provenance verification.

\subsection{A borderline set cannot be inherited}
\label{sec:transfer}

The seven items of Section~\ref{sec:empirical} were constructed against
\texttt{gpt-4o}'s decision boundary.  Graded with no sampling
parameter sent by Claude Opus~5.5 and Claude Opus~4.8 on 28~October,
across 700~calls at $N{=}50$ per item per model, not one item flipped,
and the two models returned identical verdicts on all seven.  Pooling
with a prior $N{=}10$ run of the same date, which
Table~\ref{tab:entropy} does not list, gives 60~runs per item per model with no
flip, which bounds the per-item flip rate at about 5\% at 95\%
confidence by the rule of three.

On 12~November I ran the same items again, with the same prompt and
the same extraction, at $N{=}50$ on Claude Opus~4.6, Opus~4.7 and
Opus~4.8: 1{,}400 calls, none of which failed.  Opus~5.5 was not
re-run.  Opus~4.8 returned one dissenting verdict: item~3 came back C
on forty-nine runs and I on one, where all sixty earlier runs had been
C.  That is inside the bound just stated, and it is why the October
result was stated as a bound and not as a zero.  Opus~4.7 split
item~6, 38~C against 12~I.  Opus~4.6 split items~5 and~6, 49--1 and
48--2; sent \texttt{temperature=0}, which that model still accepts
under the older API version, it split the same two items 49--1 and
49--1.  The two dates differ in the API key used as well as in time,
and I do not attribute the difference on Opus~4.8 to either.

Neither run is evidence that the current models are stable graders,
and the November run shows as much directly for Opus~4.8.  What the two
runs show together is that these items are far from the boundary of
Opus~5.5 and Opus~4.8: across both dates the two depart
from unanimity on one verdict in 1{,}050 at $N{=}50$, on items that
\texttt{gpt-4o} splits four of seven.  The set carries over in part to
Opus~4.6 and Opus~4.7 and not in a way that could have been predicted from
\texttt{gpt-4o}: item~6, which \texttt{gpt-4o} splits, also splits on
Opus~4.6 and Opus~4.7, while item~4, the one \texttt{gpt-4o} leaves
nearest an even split, is returned C on every run by all four Anthropic
models.  A calibration set
is a property of the grader it was built against, and cannot be
inherited by a successor.  The practical consequence for a harness is
that the boundary has to be re-derived whenever the grader changes,
which is an operating cost the field has not been paying.

\begin{table}[h]
\centering\small
\caption{Majority verdict on each of the seven items, by grader, at
  $N{=}50$ with no sampling parameter sent, 28~October~2026; the OpenAI
  columns are the chat completions endpoint, and their heads omit the
  \texttt{gpt-} prefix.  A count below 50
  is set in bold: the grader did not return the same verdict every
  time.  In the 12~November repeat Opus~4.8 returned C\,49 on item~3.  Items~1, 5 and~6 receive opposite verdicts from different
  graders, and in several of those cells the disagreeing graders are
  individually unanimous.}
\label{tab:verdicts}
\begin{tabular}{@{}l*{7}{c}@{}}
\toprule
Item & Opus 5.5 & Opus 4.8 & \texttt{5.5} & \texttt{5.6-sol} &
\texttt{6-sol} & \texttt{6.1-sol} & \texttt{6.1-terra} \\
\midrule
1 & I\,50 & I\,50 & \textbf{C}\,46 & \textbf{C}\,27 & I\,50 & I\,50 & I\,50 \\
2 & I\,50 & I\,50 & I\,50 & I\,50 & I\,50 & I\,50 & I\,50 \\
3 & C\,50 & C\,50 & C\,50 & C\,50 & C\,50 & C\,50 & C\,50 \\
4 & C\,50 & C\,50 & C\,50 & C\,50 & C\,50 & C\,50 & C\,50 \\
5 & C\,50 & C\,50 & C\,50 & \textbf{C}\,44 & \textbf{I}\,42 & \textbf{C}\,46 & I\,50 \\
6 & C\,50 & C\,50 & C\,50 & C\,50 & C\,50 & \textbf{C}\,48 & \textbf{I}\,34 \\
7 & C\,50 & C\,50 & C\,50 & C\,50 & C\,50 & C\,50 & C\,50 \\
\bottomrule
\end{tabular}
\end{table}

\subsection{The verdict is not stable across graders}
\label{sec:verdicts}

Section~\ref{sec:transfer} reports that the inherited items barely
split on the current Anthropic models.  Graded by five current OpenAI
models at $N{=}50$ on the chat completions endpoint, again with no
sampling parameter sent, they do split, and
they do something the within-grader statistics cannot show.
Table~\ref{tab:verdicts} gives the majority verdict of each grader on
each item.

Three of the seven items receive opposite verdicts from different
graders.  Item~5 is graded C by \texttt{gpt-5.5} on all fifty runs
and by \texttt{gpt-5.6-sol} on forty-four, I by \texttt{gpt-6-sol} on
forty-two, C again by \texttt{gpt-6.1-sol} on forty-six, and I by
\texttt{gpt-6.1-terra} on all fifty: the verdict reverses three times
across five models of one vendor, and the first and the last of them
never once wavered.  Item~1 divides the
vendors, C on \texttt{gpt-5.5} and I on both Anthropic models and on
every OpenAI model from \texttt{gpt-6} onward.  Item~6 is C on every
grader but \texttt{gpt-6.1-terra}.

The consequence for a harness is direct and is not a variance problem.
An evaluation that upgrades its grader, which is the ordinary
consequence of following a provider's current model, can move its
pass/fail line while measuring zero disagreement before the move and
zero after it.  Both graders report a stable verdict; the verdicts are
opposite.  No amount of repeated sampling within either grader
discloses this, because within either grader there is nothing to
disclose.  It follows that a disagreement metric computed against one
grader is not a property of the evaluation, and that a harness should
record which grader produced a verdict with the same care it records
the verdict.

\subsection{A boundary set, obtained rather than inherited}
\label{sec:boundary}

If the inherited items are not near the current grader's boundary, the
question is whether that boundary can be found.  I generated 2{,}001
candidate items with a small model, each built so that a strict and a
generous reading of the criterion disagree, spanning twenty domains and
ten ways for a submission to fall short.  I graded all 2{,}001 of
them four times each on Claude Opus~5.5 at the enforced setting, the
first 1{,}000 on 28~October and the rest on 31~October: 8{,}004 calls,
of which 46~items, 2.3~per cent, did not come back unanimous (20 of
the first 1{,}000 and 26 of the remaining 1{,}001).  I then ran each of
the 46 fifty times.  Forty-five were still split; one, on the Hawthorne
effect, came back I on all fifty, so the four-run screen produced one
false positive in 46.  Runs that returned no verdict are excluded from
the run count $n$: 259 of the 2{,}250 confirmation runs of the
forty-five.  On the later 26 items, where the stop reason was
recorded, all 183 such runs had spent the 1{,}024-token output budget
on reasoning before reaching a verdict.  On seven items fewer than
forty valid runs remain.

Table~\ref{tab:boundary} gives the forty-five.  The mean is
$H=0.763$, against $\bar{H}=0.000$ for the inherited items on the same
grader over 350~calls; twenty-three of the items are above $H=0.9$,
and two sit at $H=1.000$, the item on creep returning 21~C against
21~I and the item on the Hardy--Weinberg assumptions 24 against 25: a
judge retaining one bit of noise and no information, on a current
frontier model, at the only setting its caller is permitted.

\begin{table}[tp]
\centering\footnotesize
\caption{A boundary set for Claude Opus~5.5, obtained by search rather
  than inherited.  Every item here was still split after fifty runs;
  $n$ counts the runs that returned a verdict.  Mean $H=0.763$, against
  $\bar{H}=0.000$ for the seven inherited items on the same grader over
  350~calls.  The one four-run survivor that resolved, on the Hawthorne
  effect, is omitted.}
\label{tab:boundary}
\begin{tabular}{@{}llrcr@{}}
\toprule
Domain & Item & $n$ & C\,/\,I & $H$ \\
\midrule
materials science & What is creep, and what are its characteristic\,\ldots & 42 & 21\,/\,21 & 1.000 \\
biology & List the assumptions of the Hardy-Weinberg equ\,\ldots & 49 & 24\,/\,25 & 1.000 \\
psychology & According to attachment theory, why do infants\,\ldots & 48 & 25\,/\,23 & 0.999 \\
history & What is the term for the grant of land held in\,\ldots & 40 & 19\,/\,21 & 0.998 \\
history & What was the main purpose of the Domesday Book\,\ldots & 46 & 25\,/\,21 & 0.995 \\
medicine & What is the mechanism of the flushing and hypo\,\ldots & 50 & 22\,/\,28 & 0.990 \\
ecology & What is the difference between a fundamental n\,\ldots & 41 & 23\,/\,18 & 0.989 \\
materials science & Explain the mechanism of the shape memory effe\,\ldots & 30 & 13\,/\,17 & 0.987 \\
law & When must police give Miranda warnings before\,\ldots & 30 & 13\,/\,17 & 0.987 \\
cryptography & What does a message authentication code provid\,\ldots & 37 & 21\,/\,16 & 0.987 \\
music theory & Name the chord built on the lowered second sca\,\ldots & 49 & 21\,/\,28 & 0.985 \\
economics & Define opportunity cost. & 50 & 29\,/\,21 & 0.981 \\
history & Which Mughal emperor commissioned the Taj Maha\,\ldots & 44 & 26\,/\,18 & 0.976 \\
logic & Why is first-order induction weaker than secon\,\ldots & 49 & 20\,/\,29 & 0.976 \\
geography & Why is Antarctica classified as a desert? & 50 & 20\,/\,30 & 0.971 \\
biology & What happens to red blood cells placed in diff\,\ldots & 49 & 30\,/\,19 & 0.963 \\
psychology & What is the difference between reliability and\,\ldots & 50 & 31\,/\,19 & 0.958 \\
economics & Why does expansionary monetary policy in an op\,\ldots & 37 & 14\,/\,23 & 0.957 \\
chemistry & Why does aluminum corrode very little in air e\,\ldots & 44 & 28\,/\,16 & 0.946 \\
psychology & What is a confound in an experimental design? & 44 & 16\,/\,28 & 0.946 \\
medicine & What is the specific reversal agent for dabiga\,\ldots & 46 & 16\,/\,30 & 0.932 \\
physics & Why does the pressure of a gas rise when it is\,\ldots & 40 & 13\,/\,27 & 0.910 \\
geography & Why are the Himalayas still rising today? & 44 & 30\,/\,14 & 0.902 \\
psychology & How does metacognition help explain the Dunnin\,\ldots & 41 & 11\,/\,30 & 0.839 \\
physics & Why is the period of a simple pendulum nearly\,\ldots & 42 & 11\,/\,31 & 0.830 \\
linguistics & Why is the English negative prefix in 'unhappy\,\ldots & 48 & 37\,/\,11 & 0.777 \\
ecology & According to island biogeography theory, what\,\ldots & 31 & 24\,/\,7 & 0.771 \\
geography & Why do rivers form meanders? & 47 & 37\,/\,10 & 0.747 \\
history & What was George Kennan's contribution to the C\,\ldots & 47 & 37\,/\,10 & 0.747 \\
economics & Why can two countries gain from trade even whe\,\ldots & 44 & 36\,/\,8 & 0.684 \\
history & What did the Taiping Rebellion achieve and how\,\ldots & 50 & 9\,/\,41 & 0.680 \\
history & What economic effect did the Black Death have\,\ldots & 50 & 42\,/\,8 & 0.634 \\
ecology & What distinguishes an invasive species from an\,\ldots & 47 & 7\,/\,40 & 0.607 \\
cryptography & Why is a deliberately slow key-derivation func\,\ldots & 49 & 42\,/\,7 & 0.592 \\
computer networking & What does the TTL of a DNS resource record con\,\ldots & 47 & 41\,/\,6 & 0.551 \\
materials science & Why does rapidly quenched carbon steel become\,\ldots & 50 & 44\,/\,6 & 0.529 \\
statistics & How does survivorship bias distort conclusions\,\ldots & 37 & 34\,/\,3 & 0.406 \\
computer networking & What does the MTU of a link define? & 50 & 4\,/\,46 & 0.402 \\
law & What must a claimant prove to establish liabil\,\ldots & 40 & 3\,/\,37 & 0.384 \\
music theory & Two voices move in similar motion into a perfe\,\ldots & 41 & 38\,/\,3 & 0.378 \\
computer networking & Why does a host behind a NAT device usually fa\,\ldots & 28 & 26\,/\,2 & 0.371 \\
history & Why did the Janissary corps decline as a milit\,\ldots & 47 & 44\,/\,3 & 0.342 \\
operating systems & What is thrashing? & 46 & 2\,/\,44 & 0.258 \\
computer networking & What is the purpose of a DHCP lease? & 50 & 2\,/\,48 & 0.242 \\
logic & Is validity in first-order logic decidable? & 50 & 2\,/\,48 & 0.242 \\
\bottomrule
\end{tabular}
\end{table}

\subsubsection*{The set does not carry to another grader either}

Section~\ref{sec:transfer} reports one direction: items built against
\texttt{gpt-4o} are returned unanimously, or within one verdict of it,
by the current Anthropic models.
That direction alone admits a reading in which the later model is
simply better and the older items are simply easy.  The boundary set
obtained above lets the question be put the other way.  I graded all
forty-five confirmed items on two current OpenAI models at $N{=}50$
on the chat completions endpoint with no sampling parameter sent,
4{,}500 calls in all; Table~\ref{tab:cross} gives the result.  For
\texttt{gpt-6.1-sol} neither endpoint admitted \texttt{temperature=0}.
For \texttt{gpt-5.6-sol} it is the default of that endpoint and not an
enforced setting, since Section~\ref{sec:withdrawal} records a route to
\texttt{temperature=0} with reasoning switched off; I did not grade
under that route, and the \texttt{gpt-5.6-sol} column should be read as
the default condition.

They carry poorly, and unevenly.  \texttt{gpt-6.1-sol} returns thirty-nine of the
forty-five unanimously, with a mean of $H=0.068$ against $0.763$ on the
grader the items were built against; \texttt{gpt-5.6-sol} returns
twenty-four unanimously, with a mean of $0.280$.  Twenty-two of the
forty-five are unanimous on both OpenAI graders at fifty runs, and nine
of those twenty-two are items Opus~5.5 leaves above $H=0.9$.  Among
the nine is the Hardy--Weinberg item that Opus~5.5 splits 24 against
25, $H=1.000$ to three places: both OpenAI graders return it C on all
fifty runs.
An item that one grader cannot settle in fifty attempts is, for
another grader, not a hard item at all.

The reading that the later model is merely better does not survive the
second column.  \texttt{gpt-5.6-sol} is the earlier of the two OpenAI
models and retains twenty-one of the forty-five, while
\texttt{gpt-6.1-sol} retains six; only four items are retained by both,
and two are retained by \texttt{gpt-6.1-sol} alone.  The same two
models differed in the same direction on the inherited seven items of
Section~\ref{sec:empirical}.

Three of the forty-five are returned unanimously by both OpenAI
graders and in opposite directions.  On the ecology item asking for the
difference between a fundamental and a realized niche, which Opus~5.5
splits 23--18 at $H=0.989$, \texttt{gpt-5.6-sol} returns C on all fifty
runs and \texttt{gpt-6.1-sol} returns I on all fifty.  On the
music-theory item on two voices moving in similar motion into a perfect
fifth, and on the logic item asking whether validity in first-order
logic is decidable, the directions are the other way round:
\texttt{gpt-5.6-sol} returns I fifty times and \texttt{gpt-6.1-sol}
returns C fifty times.  Neither grader wavers on any of the three, and
the two of them return opposite verdicts.  An entropy of zero is
therefore not evidence that a grader has the item right; it is evidence
that the grader will keep saying the same thing.  A harness that
watched only the spread of its own grader's verdicts would find nothing
to report in either column.

Where a grader's boundary lies is a property of that grader, not a
position on a scale of difficulty along which models advance.  A
calibration set therefore cannot be inherited across vendors, across
generations, or between sibling models of one generation, and the
procedure that produces one is worth more than any particular set it
produces.

\begin{table}[tp]
\centering\footnotesize
\caption{The boundary set of one grader, put to two others at
  $N{=}50$.  Entries are the per-item normalized entropy; a dash marks
  an item the grader returned unanimously.  The twenty-three items
  listed are those that are not unanimous on at least one OpenAI
  grader; the remaining twenty-two of the forty-five are unanimous on
  both and are omitted.  The summary rows are over all forty-five.
  These items were obtained by searching for Claude Opus~5.5's
  boundary; nearly all of them are unanimous for \texttt{gpt-6.1-sol},
  and about half are unanimous for \texttt{gpt-5.6-sol}.}
\label{tab:cross}
\begin{tabular}{@{}lccc@{}}
\toprule
Item & Opus 5.5 & \texttt{gpt-5.6-sol} & \texttt{gpt-6.1-sol} \\
\midrule
What is creep, and what are its characteri\,\ldots & 1.000 & 0.141 & \,-- \\
According to attachment theory, why do inf\,\ldots & 0.999 & 0.327 & \,-- \\
What is the term for the grant of land hel\,\ldots & 0.998 & \,-- & 0.141 \\
What was the main purpose of the Domesday\,\ldots & 0.995 & 0.242 & \,-- \\
Explain the mechanism of the shape memory\,\ldots & 0.987 & 0.904 & 0.634 \\
When must police give Miranda warnings bef\,\ldots & 0.987 & 0.990 & 0.795 \\
Define opportunity cost. & 0.981 & 0.141 & \,-- \\
What happens to red blood cells placed in\,\ldots & 0.963 & 0.958 & \,-- \\
Why does expansionary monetary policy in a\,\ldots & 0.957 & 0.881 & \,-- \\
Why does aluminum corrode very little in a\,\ldots & 0.946 & 0.855 & \,-- \\
What is a confound in an experimental desi\,\ldots & 0.946 & 0.760 & \,-- \\
What is the specific reversal agent for da\,\ldots & 0.932 & 0.242 & \,-- \\
Why does the pressure of a gas rise when i\,\ldots & 0.910 & 0.402 & \,-- \\
Why are the Himalayas still rising today? & 0.902 & 0.943 & \,-- \\
How does metacognition help explain the Du\,\ldots & 0.839 & 0.529 & \,-- \\
Why is the English negative prefix in\,\ldots & 0.777 & 0.760 & 1.000 \\
What was George Kennan's contribution to t\,\ldots & 0.747 & 0.680 & \,-- \\
Why can two countries gain from trade even\,\ldots & 0.684 & 0.827 & 0.242 \\
What did the Taiping Rebellion achieve and\,\ldots & 0.680 & 0.981 & \,-- \\
Why is a deliberately slow key-derivation\,\ldots & 0.592 & 0.242 & \,-- \\
What does the TTL of a DNS resource record\,\ldots & 0.551 & 0.402 & \,-- \\
Why does a host behind a NAT device usuall\,\ldots & 0.371 & \,-- & 0.242 \\
Why did the Janissary corps decline as a m\,\ldots & 0.342 & 0.402 & \,-- \\
\midrule
Split items, of 45 & 45 & 21 & 6 \\
Mean $H$ & 0.763 & 0.280 & 0.068 \\
\bottomrule
\end{tabular}
\end{table}

\subsubsection*{The same search on the other grader}

The screen is cheap enough to run on the other vendor, and running it
there answers the question Table~\ref{tab:cross} leaves open: whether
\texttt{gpt-6.1-sol} has a boundary of its own in this pool, or is
simply unanimous on it.  I put all 2{,}001 candidates to
\texttt{gpt-6.1-sol} four times each on the chat completions endpoint
with no sampling parameter sent; this model admitted
\texttt{temperature=0} on neither endpoint.
The OpenAI credit ran out at the tail of the pool: 115 of the 8{,}004
calls returned a no-credit error, all of them on the last 31 candidates
in pool order, of which 25 received no valid grade and six received
fewer than four.  Because the screen appends each verdict as it
returns and walks the pool in order, the loss is confined to those 31
items; no item elsewhere is short a run, and no survivor is counted on
fewer than four valid grades.  I therefore report the screen over the
1{,}970 candidates that completed and treat the 31 as unscreened rather
than as unanimous.  Of the 1{,}970, 21 did not come back unanimous,
1.1~per cent; fifteen split 3--1 and six split 2--2
(Table~\ref{tab:boundary_oai}).  These are four-run survivors.  The
Opus set above was confirmed at higher run counts before being reported
as a boundary, and the same confirmation for these 21 is scheduled and
not yet done, so the table reports the four-run split and not a settled
entropy.

The two searches ran over the same pool of 2{,}001 candidates, 1{,}970
of which completed the OpenAI screen, so the sets
can be compared directly.  Claude Opus~5.5 retained 46 and
\texttt{gpt-6.1-sol} retained 21; three items are in both, the
geography item asking why Antarctica is classified as a desert, the
linguistics item on the negative prefix in \emph{unhappy}, and the
history item on land held for military service.  The Jaccard overlap
of the two sets is 0.047.  One of the three did not hold its split on
\texttt{gpt-6.1-sol}: the Antarctica item, 2--2 in the four-run screen
of October, was returned I on all fifty runs by the same model on
1~November (it is among the items Table~\ref{tab:cross} omits as
unanimous).  A four-run split on this grader is therefore not yet a
boundary item, which is why the 21 are reported as survivors of a
screen and not as a confirmed set.  The domains differ as well: the Opus set is
led by history, psychology and computer networking and spans eighteen
domains; the OpenAI set is led by computer networking and operating
systems.  This is the other half of the result Table~\ref{tab:cross}
gives in one direction.  The items that survive the screen on one
grader are, with three exceptions, unanimous for the other at four
runs, and conversely; the search turns up split items on each grader,
at yields of 2.3 and 1.1~per cent, and the two sets of survivors are
nearly disjoint.  Only the Opus set has been confirmed at fifty runs.

\begin{table}[h]
\centering\small
\caption{Four-run survivors for \texttt{gpt-6.1-sol}, from the same
  search over the same pool: candidates for a boundary set, not yet
  confirmed at fifty runs.  Entries are the C\,/\,I split at four
  runs; the 2--2 items are listed first.  Of 2{,}001 candidates, 1{,}970
  completed four runs and 21 were not unanimous.  Three items are
  shared with the Opus set of Table~\ref{tab:boundary}: the geography
  item on Antarctica, the linguistics item on the negative prefix, and
  the history item on land held for military service.}
\label{tab:boundary_oai}
\begin{tabular}{@{}llc@{}}
\toprule
Domain & Item & C\,/\,I \\
\midrule
statistics & Why do extreme first measurements tend to be l\,\ldots & 2\,/\,2 \\
geography & Why is Antarctica classified as a desert? & 2\,/\,2 \\
computer networking & Why can a stateful firewall admit reply traffi\,\ldots & 2\,/\,2 \\
linguistics & Why is the English negative prefix in 'unhappy\,\ldots & 2\,/\,2 \\
linguistics & State the direction of the first consonant shi\,\ldots & 2\,/\,2 \\
operating systems & In a preemptive scheduler, how does the operat\,\ldots & 2\,/\,2 \\
computer networking & What is a half-open TCP connection? & 3\,/\,1 \\
operating systems & What is a deadlock in an operating system? & 1\,/\,3 \\
law & What must be shown for a finding of anticompet\,\ldots & 1\,/\,3 \\
medicine & Why is a test's sensitivity independent of dis\,\ldots & 3\,/\,1 \\
ecology & Why do predator populations often peak after t\,\ldots & 1\,/\,3 \\
computer networking & Why does a NAT device typically use port trans\,\ldots & 3\,/\,1 \\
computer networking & Why does DHCP assign addresses as time-limited\,\ldots & 1\,/\,3 \\
history & What is the term for the grant of land held in\,\ldots & 1\,/\,3 \\
logic & In the normal modal logic K, what is the distr\,\ldots & 3\,/\,1 \\
operating systems & What is the usable capacity of a RAID 5 array,\,\ldots & 1\,/\,3 \\
ecology & What is allelopathy? & 1\,/\,3 \\
physics & Why does the Moon always present the same face\,\ldots & 1\,/\,3 \\
mathematics & State what it means for the series sum a\_n to\,\ldots & 3\,/\,1 \\
psychology & State the encoding specificity principle. & 3\,/\,1 \\
physics & Why does the daytime sky appear blue? & 3\,/\,1 \\
\bottomrule
\end{tabular}
\end{table}

This is the artefact the transfer argument requires.  A harness cannot
inherit a calibration set, but it can obtain one, and the cost of
obtaining each of these was eight thousand calls of screening at a
yield of one to two and a half per cent, and, for the Opus set, fifty
confirmation runs on each survivor.  The procedure is the contribution, not the forty-five items
or the twenty-one, which are a property of one
grader at one date and will need deriving again at the next.

\subsection{Limitations}

The 7-item protocol is an adversarial stress test, not a
representative sample.  I demonstrate that non-determinism
\emph{can} persist, not that it \emph{typically} does at a
particular rate across all evaluation items.  Estimating the
base rate would require access to a large, representative
evaluation corpus and is left as future work.

The June runs use $N{=}10$ to $20$ per item, which is sufficient for
qualitative detection but too small for precise per-item disagreement
estimates; the seven-item, confirmation and cross-grading runs of
October and November use $N{=}50$, apart from one $N{=}10$ run on the
two Opus models, and the screens use four.  I state bounds where
relevant and do not overclaim.

The later measurements have bounds of their own.  The map of
Section~\ref{sec:withdrawal} is one request per model, endpoint and
setting from two accounts on stated dates, and it records whether a
request was accepted, not what temperature the provider then applied.
I did not grade under the one route that still admits
\texttt{temperature=0} on four current OpenAI models, the Responses
endpoint with reasoning switched off, so nothing here says how those
graders behave under it.  Claude Opus~5.5 was not re-run in November.
The boundary set for Claude Opus~5.5 is confirmed at fifty runs; the
set for \texttt{gpt-6.1-sol} is a four-run screen and has not been
confirmed.  The candidate items were generated by a model, and both
sets are properties of one grader at one date.  Each vendor's
top-tier model is outside the October map and every grading run, as
stated in Section~\ref{sec:withdrawal}.

The empirical observation in \S\ref{sec:transmission} carries
its own bounds.  The 11-day window captures all Alibaba-ecosystem
accesses on a single day (2026-06-12) with no recurrence, but
cannot establish a pre-deployment baseline.  The five accesses
are attributed by ASN (Tier~2 evidence); ASN attribution does
not identify the originating tenant, so the appropriate referent
is ``Alibaba-ecosystem ASN'' or ``Alibaba Cloud customer,'' not
the vendor as a corporate actor.  Only one frontier model
(\texttt{qwen-plus}) was probed; other tiers and other
Chinese-hosted endpoints are left as future work.  The SemrushBot
TDM comparison is from the same tracker and window but involves
a single SEO vendor; generalisation to other commercial crawlers
requires broader observation.  These bounds weaken any claim of
``sustained crawling'' but do not weaken the
capability-versus-execution separation that the section actually
makes.

\subsection{Relation to prior work on judge temperature}
\label{sec:relwork}

Two bodies of work bear directly on this, and I state the boundary
rather than leave it to the reader.  Atil et al.\
\citep{atil2024stability} establish the phenomenon: across five models
and eight tasks, output under nominally deterministic settings is
rarely reproducible, with accuracy varying by up to fifteen points
between runs.  Li et al.\ \citep{li2026temperature} carry the question
into LLM-as-judge and settle its practical form over more than a
million evaluations: lower temperatures give more consistent verdicts
and fewer errors, the effect concentrates on ambiguous items, and
repeated sampling at a higher temperature exposes that ambiguity, so
that majority voting, or high-temperature reasoning with a
low-temperature final decision, raises agreement where it matters.  I
take all of this as established and do not re-measure it.

This paper begins where the knob ends.  Every recommendation in that
line is an instruction to set a value, and
Section~\ref{sec:withdrawal} reports how little of it is left on the
current models of both vendors: Anthropic rejects
\texttt{temperature=0} from
Opus~4.7 onward, and on every model under its newer API version, and
OpenAI's chat completions endpoint pins it to its default from
\texttt{gpt-5.5} onward, having restored it for three releases in
between.  A rule that says to keep the temperature low and to raise it
when the judgment is hard is not wrong; as of November~2026 it cannot
be executed on Anthropic's frontier or on \texttt{gpt-6.1-sol}, on the
other current OpenAI models only its lower end can be requested, on one
endpoint and with reasoning switched off, and no deprecation schedule
says when that will change.  Two consequences follow, which the earlier
work does not address because it did not need to.  What must be
reported is no longer a chosen setting and its variance but the spread
produced by whatever setting the provider leaves open, which is why
Section~\ref{sec:mitigations} gives an index rather than a recommended
value.  And the instability that gates a decision is not only within a
grader: Section~\ref{sec:verdicts} shows one item receiving opposite
verdicts from two models of a single vendor, each on all fifty runs,
and Section~\ref{sec:boundary} shows three more, so a harness that upgrades its judge can move its
pass/fail line with zero measured disagreement on either side of the
move.  Since a borderline set is a property of the grader it was built
against, calibration cannot be inherited either;
Section~\ref{sec:boundary} derives one for a current model and reports
it.

\section{Mitigations}
\label{sec:mitigations}

\subsection{An index that survives the withdrawal}

With the sampler out of the evaluator's hands, what remains reportable
is the spread of the grader's own verdicts.  For a binary grade over
$n$ runs of one item, let $H$ be the Shannon entropy of the observed
C/I distribution in bits.  A binary grade carries at most one bit, so
$H$ is already on $[0,1]$ and needs no further normalization: $H=0$ is
a unanimous item and $H=1$ is an item on which the judge has retained
nothing.  I report the per-item $H$ and the condition mean
$\bar{H}$.

The index has three properties that matter at the sample sizes an
evaluator can afford.  It requires no rate estimate, which is what
Bowyer et~al.\ \citep{bowyer2025clt} warn against at these sizes: on
small benchmarks, central-limit intervals understate uncertainty and
produce error bars that are too narrow.  It is defined at $n=2$.  And
it orders conditions even where every per-item proportion has an
interval too wide to separate them.  Table~\ref{tab:entropy} gives the
index across my conditions.

\begin{table}[h]
\centering\small
\caption{Grader instability by condition, measured as the mean per-item
  normalized entropy $\bar{H}$ of the grade distribution. A binary grade
  carries at most one bit, so $\bar{H}$ needs no further normalization.
  Computed from the per-call records.  ``No parameter sent'' means that
  the call carried no sampling parameter: Opus~4.7, 4.8 and~5.5
  reject \texttt{temperature=0}, and for the OpenAI rows this is the default of
  the chat completions endpoint.  Rows dated
  12~Nov are the November repeat; the other $N{=}50$ rows are
  28~October.}
\label{tab:entropy}
\begin{tabular}{@{}llrrcr@{}}
\toprule
Grader & Condition & $N$ & Calls & Flipping items & $\bar{H}$ \\
\midrule
gpt-4o & default (temperature unset) (run~1) & 20 & 140 & 4/7 & 0.362 \\
gpt-4o-2024-08-06 & \texttt{temperature=0} (run~2) & 10 & 70 & 2/7 & 0.282 \\
gpt-4o-2024-08-06 & \texttt{temperature=0}, \texttt{top\_p}=0.1 (run~2) & 10 & 70 & 2/7 & 0.282 \\
claude-sonnet-4-6 & default (temperature unset) (run~1) & 20 & 140 & 2/7 & 0.242 \\
gpt-4o-2024-08-06 & \texttt{temperature=0}, \texttt{top\_p}=0.5 (run~2) & 10 & 70 & 2/7 & 0.229 \\
\texttt{gpt-5.6-sol} & no parameter sent & 50 & 350 & 2/7 & 0.218 \\
gpt-4o & \texttt{temperature=0} (run~1) & 10 & 70 & 2/7 & 0.206 \\
claude-sonnet-4-6 & \texttt{temperature=0} (run~1) & 10 & 70 & 2/7 & 0.170 \\
\texttt{gpt-6.1-terra} & no parameter sent & 50 & 350 & 1/7 & 0.129 \\
claude-opus-4-7 & no parameter sent (12~Nov) & 50 & 350 & 1/7 & 0.114 \\
claude-sonnet-4-6 & \texttt{temperature=0}, \texttt{top\_k}=1 (run~2) & 10 & 70 & 1/7 & 0.103 \\
\texttt{gpt-6.1-sol} & no parameter sent & 50 & 350 & 2/7 & 0.092 \\
\texttt{gpt-6-sol} & no parameter sent & 50 & 350 & 1/7 & 0.091 \\
claude-sonnet-4-6 & \texttt{temperature=0} (run~2) & 10 & 70 & 1/7 & 0.067 \\
claude-haiku-4-5-20251001 & \texttt{temperature=0} (run~2) & 10 & 70 & 1/7 & 0.067 \\
\texttt{gpt-5.5} & no parameter sent & 50 & 350 & 1/7 & 0.057 \\
claude-opus-4-6 & no parameter sent (12~Nov) & 50 & 350 & 2/7 & 0.055 \\
claude-opus-4-6 & \texttt{temperature=0} (12~Nov) & 50 & 350 & 2/7 & 0.040 \\
claude-opus-4-8 & no parameter sent (12~Nov) & 50 & 350 & 1/7 & 0.020 \\
claude-opus-5-5 & no parameter sent & 50 & 350 & 0/7 & 0.000 \\
claude-opus-4-8 & no parameter sent & 50 & 350 & 0/7 & 0.000 \\
claude-opus-4-8 & \texttt{temperature=0} (run~2) & 10 & 70 & \multicolumn{2}{c}{all 70 calls rejected} \\
\bottomrule
\end{tabular}
\end{table}

Two items reach $H=1.000$ under \texttt{temperature=0} on
\texttt{gpt-4o-2024-08-06} in run~2, item~4 in the condition without
\texttt{top\_p} and item~6 with \texttt{top\_p}=0.1: a five-five split,
a judge that has returned one bit of noise and no information.
(Table~\ref{tab:openai} shows run~1, where the same two items split
6--4 and 9--1.)  Restricting the nucleus does not help;
\texttt{top\_p} moves $\bar{H}$ by less than the spread between two
runs of the same condition.  The comparison that carries the argument is between the
rows in which no sampling parameter was sent.  They are the setting a
harness gets when it configures nothing, and they do not behave alike.
Claude Opus~5.5 sits at $\bar{H}=0.000$, and Claude Opus~4.8 at
$0.000$ in October and $0.020$ in November.  \texttt{gpt-5.6-sol},
graded on the same seven items at the chat completions default of~1,
sits at $\bar{H}=0.218$, with item~1 at $H=0.995$, 27~C against 23~I;
that same item every Opus row at $N{=}50$ returns I on all fifty
runs.  The information a judge retains at the setting its caller gets
by default is therefore a property of the provider rather than
of the evaluation, and an evaluator who changes vendor changes it
without any means of observing that they have.
Section~\ref{sec:transfer} states why $\bar{H}=0.000$ should not in
any case be read as stability.

\subsection{Reporting}

I therefore restate the recommendation of the first version of this
paper in the form the withdrawal leaves open.  A harness should run its
grader more than once per item, report $\bar{H}$ beside the score, and
record the endpoint, the API version and the sampling settings in force
at the time of the run, since those are now properties of the provider
rather than of the experiment.  Miller \citep{miller2024errorbars} gives the statistical
frame for treating an evaluation as an experiment; Lee et~al.\
\citep{lee2025reporting} correct the complementary quantity, the bias
a judge's imperfect sensitivity and specificity induce in the score.
Bias and variance both need reporting, and neither substitutes for the
other.


Based on my findings, I recommend that evaluation harnesses
adopt the following practices:

\textbf{(M1)~Set \texttt{temperature=0} explicitly} at the grader
call site, where the provider supports it.  Never rely on a
provider's \texttt{None} default.

\textbf{(M2)~Set \texttt{seed}} where supported, and record it
in run metadata.

\textbf{(M3)~Run epochs\,$>1$} for the grader and report
variance or confidence intervals, not a single point estimate.
This is the only mitigation that survives the withdrawal of
sampling control reported in Section~\ref{sec:withdrawal}, and it is
unconditional precisely because that withdrawal follows no announced
schedule (Section~\ref{sec:nottrajectory}).

\textbf{(M4)~Surface grader disagreement} as a first-class
harness health metric.  Items with high intra-grader
disagreement are candidates for rubric revision or human
adjudication.

\textbf{(M5)~Log effective grader configuration}
(model identifier, resolved version, temperature, seed,
\emph{and resolved API endpoint / base URL})
into the results artifact, so that consumers of the
evaluation know under what conditions, and on what
infrastructure, grades were produced
(\S\ref{sec:transmission}).

\section{Conclusion}
\label{sec:conclusion}

I have shown that the standard mitigation for LLM-as-judge
non-determinism (setting \texttt{temperature=0}) is necessary
but not sufficient.  In the concrete case of Japan AISI's
evaluation harness, the mitigation is not even applied: the
grader runs at temperature~1.0 by silent default.  When applied,
residual non-determinism persists across providers, model tiers,
and sampling configurations, including forced greedy decoding.

The mitigation is also no longer generally available.  As of
November~2026 Anthropic's current models reject \texttt{temperature=0},
and every model does under the newer API version; OpenAI's chat completions
endpoint pins it to the default on its current models, and the
Responses endpoint admits \texttt{temperature=0} on some of them only
with reasoning switched off, and on \texttt{gpt-6.1-sol} not at all.
The withdrawal did not follow the single deprecation trajectory the
first version of this paper predicted, which is why it cannot be
planned against.  What an evaluator can still do is run the grader
repeatedly and report the spread, and the per-item normalized entropy
is a form of that report which needs no sampler setting.

Nor can a calibration set be carried forward.  Items built against one
grader's boundary are mostly decided without hesitation by another,
the split items the same search finds on two graders barely
overlap, and two models of one vendor return opposite verdicts on the
same item, each on all fifty runs.  A low
entropy shows that a grader is consistent, not that it is right.

Beyond sampling, the API compatibility layer opens a
transmission-layer gap: a single configuration change can
redirect grader calls to endpoints in different jurisdictions,
leaving evaluation provenance opaque even when the grader
parameters are otherwise correct.

These findings argue that evaluation harnesses should treat
grader reliability as a measurement property, not a
configuration detail.  A grader-disagreement rate and the
resolved endpoint belong in the results artifact alongside
the scores, not because they make the evaluation less
trustworthy, but because their \emph{absence} does.

\section*{Data and code availability}

The June study (910~calls) and its harness are archived on
Zenodo~\citep{tamba2026nondeterminism}.  The per-call records of the
October probes of both vendors and the November recheck of Anthropic's,
the seven-item runs on
the Anthropic models of 28~October and 12~November, and the scripts
that produced them are public at
\url{https://github.com/hiroki-tamba-research/llm-judge-sampling-probe-202610}.
The remaining records behind this version are not public at the time
of writing: the seven-item runs on the five OpenAI models, the 2{,}001
candidate items, the four-run screens on both graders, the fifty-run
confirmation on Claude Opus~5.5, and the cross-grading of the
forty-five items.  A deposit of these records is in preparation.

\section*{Acknowledgements}

Eirik Botten Nicolaysen (avalyset / EcoDeco AS;
ORCID~\href{https://orcid.org/0009-0001-9188-6788}{0009-0001-9188-6788})
independently verified the non-determinism on separate infrastructure
and contributed observations on \texttt{top\_p}/\texttt{top\_k}
coverage and cross-model scoping that informed the v1.1 extension.



\begin{thebibliography}{9}

\bibitem[Zheng et~al.(2023)]{zheng2023judging}
L.~Zheng, W.-L. Chiang, Y.~Sheng, et~al.
\newblock Judging {LLM}-as-a-{J}udge with {MT}-{B}ench and {C}hatbot
  {A}rena.
\newblock \emph{Advances in Neural Information Processing Systems~36
  (NeurIPS 2023)}, 2023.
\newblock arXiv:2306.05685.

\bibitem[Wang et~al.(2023)]{wang2023fair}
P.~Wang, L.~Li, L.~Chen, et~al.
\newblock Large Language Models are not Fair Evaluators.
\newblock 2023.
\newblock arXiv:2305.17926.

\bibitem[He et~al.(2025)]{he2025defeating}
H.~He et~al.
\newblock Defeating Nondeterminism in {LLM} Inference.
\newblock Thinking Machines Lab, 2025.
\newblock \url{https://thinkingmachines.ai/blog/defeating-nondeterminism-in-llm-inference/}

\bibitem[Stureborg et~al.(2024)]{stureborg2024llmjudges}
R.~Stureborg, A.~Alikaniotis, Y.~Suhara.
\newblock Large Language Models are Inconsistent and Biased Evaluators.
\newblock 2024.
\newblock arXiv:2405.01724.

\bibitem[Shankar et~al.(2024)]{shankar2024validates}
V.~Shankar, J.~D.~Zamfirescu-Pereira, B.~Hartmann,
  A.~G.~Parameswaran, I.~Arawjo.
\newblock Who Validates the Validators?
  Aligning {LLM}-Assisted Evaluation of {LLM} Outputs with Human
  Preferences.
\newblock 2024.
\newblock arXiv:2404.12272.

\bibitem[{Japan AISI}(2026)]{aisev25}
{Japan AISI}.
\newblock aisev issue \#25: Reproducibility Improvement Proposal.
\newblock \url{https://github.com/Japan-AISI/aisev/issues/25}, 2026.

\bibitem[Liu et~al.(2026)]{liu2026agent}
H.~Liu, C.~Shou, H.~Wen, Y.~Chen, R.~J.~Fang, Y.~Feng.
\newblock Your Agent Is Mine: Measuring Malicious Intermediary
  Attacks on the {LLM} Supply Chain.
\newblock 2026.
\newblock arXiv:2604.08407.

\bibitem[Heim(2025)]{heim2025diffusion}
L.~Heim.
\newblock Understanding the Artificial Intelligence Diffusion
  Framework: Can Export Controls Create a {U.S.}-Led Global
  Artificial Intelligence Ecosystem?
\newblock {RAND} Corporation, Perspective PEA3776-1, 2025.
\newblock \url{https://www.rand.org/pubs/perspectives/PEA3776-1.html}

\bibitem[{Kong Inc.}(2026)]{kong2026claudecode}
{Kong Inc.}
\newblock Route Claude {CLI} traffic through {AI} Gateway and
  {D}ash{S}cope.
\newblock Kong Developer Documentation, 2026.
\newblock \url{https://developer.konghq.com/how-to/use-claude-code-with-ai-gateway-dashscope/}

\bibitem[Atil et~al.(2024)]{atil2024stability}
B.~Atil et~al.
\newblock Non-Determinism of ``Deterministic'' {LLM} Settings
  (originally {\emph{LLM} Stability: A Detailed Analysis with Some
  Surprises}).
\newblock 2024, revised 2025.
\newblock arXiv:2408.04667.

\bibitem[Li et~al.(2026)]{li2026temperature}
L.~Li, L.~Sleem, Y.~Xu, Y.~Song, A.~Jia, J.~Francois, R.~State.
\newblock The Necessity of Setting Temperature in {LLM}-as-a-{J}udge.
\newblock 2026, revised October 2026.
\newblock arXiv:2603.28304.

\bibitem[Miller(2024)]{miller2024errorbars}
E.~Miller.
\newblock Adding Error Bars to Evals: A Statistical Approach to
  Language Model Evaluations.
\newblock 2024.
\newblock arXiv:2411.00640.

\bibitem[Bowyer et~al.(2025)]{bowyer2025clt}
S.~Bowyer, L.~Aitchison, D.~R.~Ivanova.
\newblock Position: Don't Use the {CLT} in {LLM} Evals With Fewer Than
  a Few Hundred Datapoints.
\newblock \emph{ICML 2025} (Spotlight Position Paper), 2025.
\newblock arXiv:2503.01747.

\bibitem[Lee et~al.(2025)]{lee2025reporting}
C.~Lee, T.~Zeng, J.~Jeong, J.-y. Sohn, K.~Lee.
\newblock How to Correctly Report {LLM}-as-a-{J}udge Evaluations.
\newblock 2025, revised 2026.
\newblock arXiv:2511.21140.

\bibitem[Barry(2026)]{barry2026integer}
N.~Barry.
\newblock Eliminating Nondeterminism in {LLM} Inference via
  Integer-Only Arithmetic.
\newblock \emph{NeurIPS 2026 Workshop on On-Device Intelligence}, 2026.
\newblock \url{https://openreview.net/forum?id=yCJPh9NkW1}

\bibitem[Tamba(2026)]{tamba2026nondeterminism}
H.~Tamba.
\newblock Non-determinism in {LLM}-as-{J}udge Graders: An Empirical
  Reproducibility Note (v1.1).
\newblock Zenodo, 2026.
\newblock \href{https://doi.org/10.5281/zenodo.20674090}{doi:10.5281/zenodo.20674090}.

\end{thebibliography}
\end{document}